\documentclass[conference]{IEEEtran}
\usepackage[utf8]{inputenc}
\usepackage[T1]{fontenc}
\usepackage{amsmath,amssymb,amsfonts}
\usepackage{amsthm}
\usepackage{algorithmic}
\usepackage{graphicx}
\usepackage{textcomp}
\usepackage{xcolor}
\usepackage{hyperref}
\usepackage{booktabs}
\usepackage{listings}
\usepackage{array}
\usepackage{tikz}
\usetikzlibrary{shapes.geometric, arrows.meta, positioning, fit, backgrounds, calc}

\newtheorem{definition}{Definition}

\hypersetup{colorlinks=true,linkcolor=blue,citecolor=blue,urlcolor=blue}

\begin{document}


\title{Ethical Hyper-Velocity (EHV):\\
A Hardware-Rooted Zero-Trust Runtime Enforcement\\
Architecture for Agentic AI Systems}

\author{
\IEEEauthorblockN{Riddhi Mohan Sharma, \textit{Senior Member, IEEE}}
\IEEEauthorblockA{IAM Architect \& Non-Human Identity Governance\\
Enterprise Identity, M\&A Integration, Regulated Environments\\
Email: riddhimohan@ieee.org}
}

\maketitle


\begin{abstract}
As autonomous agentic systems scale across regulated critical infrastructures,
the lack of mechanistic, hardware-rooted enforcement for high-frequency policy
updates presents a fundamental safety gap. We present \textbf{Ethical
Hyper-Velocity (EHV)}, a governance-aware runtime enforcement architecture for
agentic systems that combines Grammar-Constrained Decoding (GCD) for inline
policy-constrained token generation, Causal Graph CRDT-based policy
synchronization with vector-clock ordering, hardware-attested execution in
Trusted Execution Environments (TEEs), and OSCAL-formatted machine-readable
audit logging. Unlike retrospective auditing frameworks (ISO/IEC 42001, NIST
AI RMF) that introduce 14--30 day policy latencies, EHV relocates the Policy
Enforcement Point (PEP) into the inference pipeline via a Governance-Aware
Just-In-Time (JIT) Compiler. Under explicitly stated assumptions, the
architecture reduces enforcement latency, improves traceability, and supports
formal verification of safety invariants in a bounded model. We demonstrate via
TLA+ model checking that non-compliant agentic actions were unreachable in the
verified bounded operating state space (1,738 states generated, 324 distinct,
depth~8, zero violations). Under these conditions, $\mathcal{O}(1)$ runtime enforcement
reduces the traditional trade-off between deployment velocity and governance
integrity, targeting Governance Latency from $\mathcal{O}(\text{days})$ toward $\mathcal{O}(1)$.
EHV's differentiating contribution is the integration of GCD, Causal CRDT,
TEE attestation caching, and bounded formal verification into a single,
hardware-rooted enforcement architecture---a combination not achieved by any
contemporaneous system. The architecture is demonstrated through a pediatric
oncology dosage use case, with applicability to regulated critical
infrastructures including healthcare, financial compliance, and critical
infrastructure control.
\end{abstract}


\section{Introduction}
\label{sec:intro}

Current AI governance relies primarily on retrospective auditing and manual
compliance gates. As autonomous agents---systems capable of multi-step reasoning
and real-world action without human intervention---proliferate in regulated
domains, this creates a \textit{Governance Bottleneck}: model execution velocity
exceeds oversight velocity by orders of magnitude.

Existing standards address the risk lifecycle (NIST AI RMF~\cite{nist_ai_rmf})
and management systems (ISO/IEC 42001~\cite{iso42001}), but they lack a
\textit{mechanistic enforcement model} at the execution layer. We argue that for
autonomous agents in high-stakes regulated environments, governance must move
from a procedural gate to a formally constrained architectural design.

We introduce \textbf{Ethical Hyper-Velocity (EHV)}, a framework for the
\textbf{runtime enforcement of AI governance policies}. EHV provides an
architecture for hardware-rooted, real-time policy enforcement that supports
safety invariants under high-frequency policy updates and distributed network
conditions. By compiling governance into the inference stack via
Grammar-Constrained Decoding, we transform policy from an external friction
point into an architecturally constrained system property.

\subsection{Contributions}
This paper makes the following contributions:
\begin{enumerate}
    \item \textbf{The Governance Latency Problem}: We formalize the metric
    $\mathit{GL} = t_{\mathrm{e}} - t_{\mathrm{d}}$ and demonstrate its measurable implications in regulated
    autonomous systems, including a 168M-action unsafe-action exposure
    calculation for a healthcare deployment.
    \item \textbf{The Identity-Action Perimeter}: We extend NIST SP~800-207
    ZTA~\cite{nist_zta} from identity verification to agentic action
    verification, closing the ``trusted identity, untrusted action'' gap via
    per-action authorization binding.
    \item \textbf{Governance-Aware JIT Compiler with GCD}: We present an
    architecture that elevates Grammar-Constrained Decoding to the primary
    enforcement path, relocating the PEP into the token-generation layer, backed
    by Causal CRDT policy synchronization and TEE-anchored epoch caching.
    \item \textbf{Bounded Formal Safety Verification}: We provide TLA+
    specifications demonstrating that non-compliant actions were unreachable in
    the verified bounded model under all explored state-space interleavings at
    depth~8.
    \item \textbf{Structured Threat Model}: We enumerate the attack surface and
    failure modes specific to governance-compiled agentic systems, including
    2026-class hardware vulnerabilities.
    \item \textbf{Regulatory Alignment}: We map EHV controls to NIST AI RMF,
    EU AI Act Article~12, NIST SP~800-207, and FDA PCCP requirements.
\end{enumerate}

\section{Related Work}
\label{sec:related}

\subsection{AI Governance Frameworks}
The NIST AI Risk Management Framework~\cite{nist_ai_rmf} provides a
lifecycle-oriented taxonomy (Map, Measure, Manage, Govern) but offers no
enforcement mechanism; compliance is assessed retrospectively. ISO/IEC
42001~\cite{iso42001} establishes a management system for AI but inherits the
PDCA audit cycle, yielding $\mathit{GL} \geq 14$ days. The EU AI Act~\cite{eu_ai_act}
mandates risk classification and conformity assessment for high-risk systems
under Annex~III, and requires automatic logging under Article~12. None of these
frameworks address real-time, pre-execution constraint enforcement for
autonomous agents executing at machine speed.

\subsection{Zero Trust Architecture}
NIST SP~800-207~\cite{nist_zta} defines Zero Trust Architecture as continuous
verification of the \textit{subject} (identity) requesting access. In agentic
contexts, this is necessary but insufficient: a cryptographically authenticated
``Physician Twin'' with valid credentials can still execute an action that
violates a policy updated seconds ago. ZTA verifies \textit{who}; EHV verifies
\textit{what}.

\subsection{Formal Methods in Safety-Critical Systems}
TLA+ and the TLC model checker~\cite{lamport_tla} have been applied to
distributed systems verification at Amazon Web Services~\cite{aws_tla} and
Microsoft Azure. Formal verification of AI safety constraints remains nascent.
Amodei et al.~\cite{amodei_concrete} enumerate concrete AI safety problems but
do not propose mechanistic enforcement. Constitutional AI~\cite{constitutional_ai}
embeds behavioral constraints in training but provides no runtime guarantee.
EHV bridges this gap by applying formal methods to the \textit{governance
enforcement layer} rather than the model itself.

\subsection{Trusted Execution Environments}
Intel SGX~\cite{intel_sgx}, AMD SEV-SNP~\cite{amd_sev}, and ARM TrustZone
provide hardware-rooted isolation for sensitive computation. Remote attestation
protocols verify enclave integrity but introduce 200ms+ latency per attestation
round-trip. EHV's Epoch-based Attestation Caching amortizes this cost to $\mathcal{O}(1)$
per inference call within an epoch.

\subsection{CRDTs for Distributed State}
Conflict-free Replicated Data Types~\cite{shapiro_crdt} guarantee eventual
consistency without coordination. Join-Semilattice structures ensure monotonic
policy convergence. EHV leverages this property to propagate safety constraints
across partitioned networks without a central policy bottleneck. The v2
architecture upgrades from Last-Write-Wins timestamps to Causal Graph CRDTs
with vector clocks~\cite{lamport_time}, removing NTP dependency and physical
clock injection attacks.

\subsection{Runtime Guardrail Systems}
NVIDIA NeMo Guardrails~\cite{nemo_guardrails} and Guardrails
AI~\cite{guardrails_ai} provide runtime constraint enforcement for LLM outputs
via programmable rules. These operate as software-layer filters between the model
and the user. EHV differs in three fundamental dimensions: (1)~\textit{formal
verification}---runtime guardrail systems offer no proof that unsafe outputs are
unreachable; (2)~\textit{hardware-rooted enforcement}---EHV's PEP executes
within a TEE, making bypass via process-level attacks significantly harder; (3)~\textit{
grammar-constrained token generation}---EHV masks invalid tokens at the logit
level before sampling, rather than filtering complete output strings.

\subsection{Grammar-Constrained Decoding Foundations}
Grammar-Constrained Decoding (GCD) compiles a context-free grammar into a DFA,
intersects the active generation prefix with allowed DFA transitions, and masks
logits for disallowed tokens to $-\infty$ before softmax. This approach is
implemented in XGrammar~\cite{xgrammar}, Outlines~\cite{outlines}, LMQL, and
vLLM's LogitsProcessor API. EHV adapts this technique specifically for policy
enforcement: the grammar encodes the complete set of permissible agentic actions
as a finite automaton, converting policy compliance into a token-generation
constraint.

\subsection{Emerging Agentic Runtime Security Systems (2025--2026)}
\label{subsec:agentic_systems}
Several contemporaneous systems address runtime governance for agentic AI:

\textbf{MI9}~\cite{mi9_2025} provides telemetry, continuous authorization, and
containment for agentic AI systems. It is the most directly comparable system in
scope---action-level governance at the application layer---but operates without
hardware TEE backing.

\textbf{AgentSpec}~\cite{agentspec_2026} offers customizable runtime enforcement
for LLM agents using declarative policy specifications. Similar in motivation but
lacks formal verification and hardware attestation.

\textbf{Aegis}~\cite{aegis_2026} mediates model outputs through a trusted
decision layer at the application tier. No hardware-rooted enforcement or formal
model checking.

\textbf{SAFi}~\cite{safi_2026} is an open-source runtime governance engine with
deterministic gating and audit logging. It provides no formal verification.

\textbf{AgenTEE}~\cite{agentee_2026} is the closest architectural parallel:
it isolates agent runtimes in TEEs using Arm CCA. However, AgenTEE does not
integrate GCD for token-level enforcement, CRDT-based policy synchronization, or
bounded formal verification.

\textbf{AutoTEE}~\cite{autotee_2026} automates porting of agent functions into
TEEs using LLM assistance, and could serve as an implementation complement to EHV.

\textit{EHV's differentiating contribution}: no existing system combines
GCD (token-level grammar enforcement), Causal CRDT (distributed policy
synchronization), TEE attestation caching, and bounded formal verification
(TLA+/TLC) in a single architecture. The contribution is not the invention of
any individual component---GCD, CRDTs, TEEs, and model checking are
independently well-established---but the \textit{co-design protocol} that
orchestrates these mechanisms into a closed enforcement loop: GCD alone does not
guarantee policy freshness (CRDT provides this), TEE alone does not constrain
token generation (GCD provides this), CRDT alone does not prevent policy
application on a compromised host (TEE attestation provides this), and formal
verification alone does not enforce at runtime (GCD+TEE provides this). This
four-way co-design creates emergent enforcement properties that no individual
component achieves.

\subsection{NIST CAISI and Agentic Identity Standards}
The NIST Center for AI Standards and Innovation (CAISI) launched the AI Agent
Standards Initiative in February 2026~\cite{nist_caisi}, addressing
interoperability, authentication, authorization, and identity management for AI
agents. The companion NCCoE concept paper on
``Accelerating the Adoption of Software and AI Agent Identity and
Authorization''~\cite{nccoe_agent_identity} proposes adapting existing identity
protocols (OAuth~2.0/2.1, OIDC, SPIFFE/SPIRE) for non-human agent workloads.
EHV's workload identity model (Section~\ref{subsec:identity}) aligns with and
supports the goals of these emerging standards.

The Model Context Protocol (MCP), donated to the Linux Foundation's Agentic AI
Foundation in December 2025~\cite{mcp_2025}, is emerging as a widely adopted
open protocol for agent-to-tool communication. EHV's action authorization layer
is compatible with MCP-structured tool calls, enabling governance enforcement at
the protocol boundary.

\section{Problem Formulation: Governance Latency}
\label{sec:problem}

\begin{definition}[Governance Latency]
For a policy decision event at time $t_{\mathrm{d}}$ and its enforcement at time $t_{\mathrm{e}}$,
Governance Latency is defined as:
\begin{equation}
\mathit{GL} = t_{\mathrm{e}} - t_{\mathrm{d}}, \quad \mathit{GL} \in [0, \infty)
\end{equation}
\end{definition}

In traditional frameworks, $\mathit{GL}$ spans 14--30 days due to manual review cycles.
During this interval, an autonomous agent may execute $N$ actions under a stale
policy state:

\begin{equation}
N_{\text{unsafe}} = \lambda \cdot \mathit{GL}
\end{equation}

where $\lambda$ is the agent's action throughput (actions/second). For
illustrative healthcare parameters---$I = 5{,}000$ Physician Twin instances,
$R = 100$ recommendations/hour---a 14-day $\mathit{GL}$ yields:

\begin{multline}
N_{\text{unsafe}} = I \cdot R \cdot 24 \cdot \mathit{GL}_{\text{days}} \\
= 5{,}000 \times 100 \times 24 \times 14 = 168{,}000{,}000
\end{multline}

This 168M-action exposure under a stale policy state establishes the
\textit{magnitude} of the Governance Latency problem. EHV's objective is to
reduce $\mathit{GL}$ toward a constant bounded by TEE attestation overhead:

\begin{equation}
\mathit{GL}_{\text{EHV}} = \mathcal{O}(1), \quad \mathit{GL}_{\text{EHV}} < 1\text{ms (target)}
\end{equation}

\textit{Note}: The $<1$ms target is an architectural design objective derived
from TEE LogitsProcessor overhead estimates. Empirical validation on production
AMD SEV-SNP hardware is identified as primary future work
(Section~\ref{sec:futurework}).

\textit{CAP Theorem Positioning}: EHV is a CP system under the CAP theorem:
it prioritizes Consistency (policy enforcement correctness) and Partition
Tolerance (CRDT convergence) over Availability. When a network partition
exceeds the epoch boundary $|E_k|$, EHV sacrifices availability by transitioning
to a fail-closed halt state (Section~\ref{sec:threats}). In safety-critical
regulated environments, halting an agent is strictly preferable to executing
under a stale or unverified policy. The $\mathcal{O}(1)$ latency target refers to
\textit{enforcement latency}---the per-token GCD check during steady-state
within a valid epoch---not end-to-end system latency including policy
propagation, which is $\mathcal{O}(\text{network})$.

\section{System Architecture}
\label{sec:architecture}

The EHV architecture consists of four technical pillars: a Causal Graph CRDT
Policy Compiler, Epoch-based Attestation Caching, a Grammar-Constrained
Decoding PEP, and a SPIFFE/SPIRE Workload Identity layer.

\subsection{Pillar 1: Causal Graph CRDT Policy Compiler}
\label{subsec:crdt}

Policies are ingested as monotonic updates in a Causal Graph CRDT using vector
clock ordering. Each policy update $p_i$ carries a vector timestamp $\vec{v}_i$.
The merge function guarantees:

\begin{equation}
\forall p_i, p_j: \vec{v}_i \succ \vec{v}_j \implies \operatorname{merge}(p_i, p_j) = p_i
\end{equation}

where $\succ$ denotes causal dominance under the vector clock partial order.
This ensures that all agent nodes eventually converge on the most recent safety
constraints without coordination overhead, and without dependence on physical
clock synchronization. The Global Ethical State $S_G$ is defined as:

\begin{equation}
S_G = \bigsqcup_{i=1}^{n} S_i
\end{equation}

where $\bigsqcup$ denotes the least upper bound in the semilattice and $S_i$
is the local state at node $i$.

\textbf{Clock Model.} The v2 specification uses vector clocks (Lamport causal
ordering~\cite{lamport_time}) in place of the v1 LWW physical timestamps. This
removes the dependency on NTP clock synchronization, preventing malicious
backdated policy injection (Threat T7, Section~\ref{subsec:threats}). In production deployments, $S_G$ is
represented as a directed acyclic graph (DAG) of cryptographically signed
policy mutations resolved via vector clocks, removing all physical clock
dependencies.

\textbf{Policy Merkle Root.} Each epoch's constraint set is committed as a
Merkle root $H_{\mathrm{p}} = \text{SHA-256}(S_G)$ enabling $\mathcal{O}(1)$ integrity verification
and GBOM binding (Section~\ref{subsec:gbom}).

\subsection{Pillar 2: Epoch-based Attestation Caching}
\label{subsec:epochs}

Remote hardware attestation (Intel TDX, AMD SEV-SNP) incurs 200ms+ latency per
round-trip. EHV introduces \textit{Policy Epochs}: the TEE validates the policy
hash $H_{\mathrm{p}}$ once per epoch $E_k$. Within an epoch, the enforcement check reduces
to:

\begin{equation}
\operatorname{Verify}(a) = 
\begin{cases}
\mathcal{O}(1), & \text{if } H_{\mathrm{p}}^{\mathrm{local}} = H_{\mathrm{p}}^{\mathrm{epoch}}, \\
\text{Re-attest}, & \text{otherwise.}
\end{cases}
\end{equation}

Epoch duration $|E_k|$ is configurable per domain. In the healthcare vertical,
$|E_k| = 60\text{s}$ balances freshness against attestation cost.

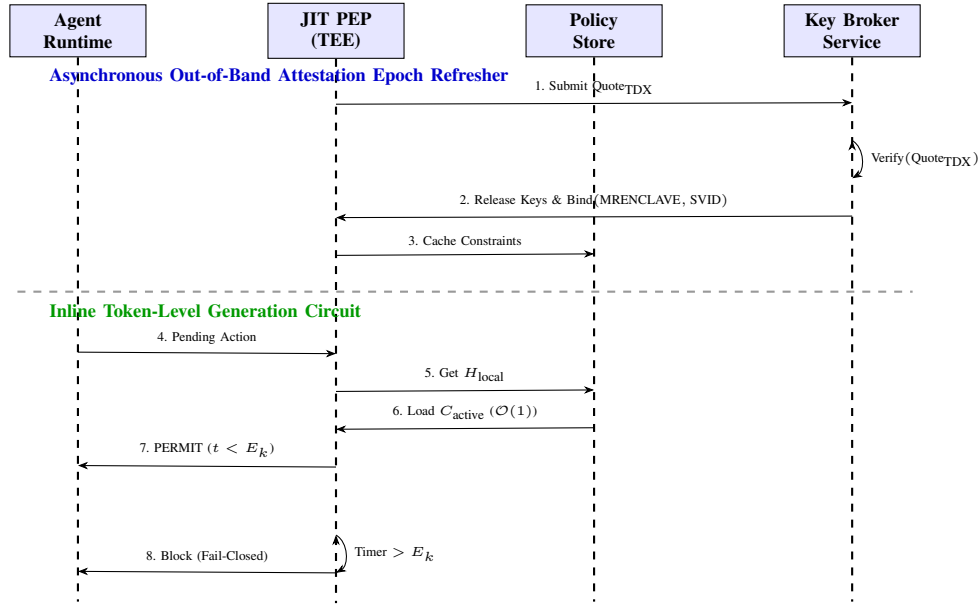
\begin{figure*}[!t]
\centering
\begin{tikzpicture}[
    node distance=2.5cm and 1.6cm,
    actor/.style={draw, rectangle, fill=blue!10, minimum width=1.8cm,
                  minimum height=0.6cm, font=\scriptsize\bfseries, align=center},
    msg/.style={-{Stealth[scale=0.8]}, font=\tiny},
    timeline/.style={thick, dashed}
]
    \node[actor] (user) {Agent\\Runtime};
    \node[actor, right=of user] (jit) {JIT PEP\\(TEE)};
    \node[actor, right=of jit] (cache) {Policy\\Store};
    \node[actor, right=of cache] (kbs) {Key Broker\\Service};

    \draw[timeline] (user.south) -- ++(0,-7.2) coordinate (u_end);
    \draw[timeline] (jit.south) -- ++(0,-7.2) coordinate (j_end);
    \draw[timeline] (cache.south) -- ++(0,-7.2) coordinate (c_end);
    \draw[timeline] (kbs.south) -- ++(0,-7.2) coordinate (k_end);

    \node[anchor=west, font=\scriptsize\bfseries\color{blue!80!black}] at ($(user.south)+(-0.5,-0.25)$) 
        {Asynchronous Out-of-Band Attestation Epoch Refresher};

    \draw[msg] ($(jit.south)+(0,-0.6)$) -- ($(kbs.south)+(0,-0.6)$)
        node[midway, above] {1. Submit $\text{Quote}_{\text{TDX}}$};
    \draw[msg] ($(kbs.south)+(0,-1.1)$) edge[bend left=60]
        node[right, font=\tiny] {$\text{Verify}(\text{Quote}_{\text{TDX}})$} ($(kbs.south)+(0,-1.6)$);
    \draw[msg] ($(kbs.south)+(0,-2.1)$) -- ($(jit.south)+(0,-2.1)$)
        node[midway, above] {2. Release Keys \& $\text{Bind}(\text{MRENCLAVE}, \text{SVID})$};
    \draw[msg] ($(jit.south)+(0,-2.6)$) -- ($(cache.south)+(0,-2.6)$)
        node[midway, above] {3. Cache Constraints};

    \draw[dashed, gray!80, thick] ($(user.south)+(-0.8,-3.1)$) -- ($(kbs.south)+(0.8,-3.1)$);

    \node[anchor=west, font=\scriptsize\bfseries\color{green!60!black}] at ($(user.south)+(-0.5,-3.35)$) 
        {Inline Token-Level Generation Circuit};

    \draw[msg] ($(user.south)+(0,-3.9)$) -- ($(jit.south)+(0,-3.9)$)
        node[midway, above] {4. Pending Action};
    \draw[msg] ($(jit.south)+(0,-4.4)$) -- ($(cache.south)+(0,-4.4)$)
        node[midway, above] {5. Get $H_{\text{local}}$};
    \draw[msg] ($(cache.south)+(0,-4.9)$) -- ($(jit.south)+(0,-4.9)$)
        node[midway, above] {6. Load $C_{\text{active}}$ ($\mathcal{O}(1)$)};
    \draw[msg] ($(jit.south)+(0,-5.4)$) -- ($(user.south)+(0,-5.4)$)
        node[midway, above] {7. PERMIT ($t < E_k$)};

    \draw[msg] ($(jit.south)+(0,-6.3)$) edge[bend left=60]
        node[right, font=\tiny] {Timer $>$ $E_k$} ($(jit.south)+(0,-6.8)$);
    \draw[msg] ($(jit.south)+(0,-6.8)$) -- ($(user.south)+(0,-6.8)$)
        node[midway, above] {8. Block (Fail-Closed)};

\end{tikzpicture}
\caption{The Core Decision Lifecycle. The JIT PEP restricts remote attestation
to periodic epoch boundaries. If a network partition exceeds the epoch
duration, the system structurally fails closed.}
\label{fig:lifecycle}
\end{figure*}

\subsection{Pillar 3: Grammar-Constrained Decoding PEP}
\label{subsec:gcd}

\textbf{v2 Primary Enforcement Path.} The v2 architecture elevates
Grammar-Constrained Decoding (GCD) to the primary Policy Enforcement Point,
replacing the v1 ASEL-based post-generation filtering. The PEP is relocated
from an external gateway to the token-generation layer of the inference pipeline.

The governance policy $\Pi$ is compiled into a Deterministic Finite Automaton
(DFA) $\mathcal{A}_\Pi$ representing the complete set of permissible action
sequences. At each generation step $t$, before softmax sampling, the GCD engine:

\begin{enumerate}
    \item Computes the current DFA state $q_t$ from the generated prefix.
    \item Queries the set of valid transitions $\Sigma_t = \mathcal{A}_\Pi.
    \text{allowed}(q_t)$.
    \item Sets logit scores for all disallowed tokens to $-\infty$:
\end{enumerate}

\begin{equation}
L^{\prime}_t[k] = 
\begin{cases}
L_t[k], & \text{if } k \in \Sigma_t, \\
-\infty, & \text{otherwise.}
\end{cases}
\end{equation}

This ensures that the model's output distribution is structurally restricted to
policy-compliant action tokens at every generation step. The constraint is
architectural, not probabilistic: no token outside $\Sigma_t$ can be sampled
regardless of its logit score.

\textbf{Double-Buffered DFA Update Protocol.} When a CRDT policy convergence
event produces an updated policy $\Pi'$ during active token generation, the
system does \textit{not} stall the inference pipeline. Instead, EHV employs a
double-buffered recompilation strategy: the updated CFG is compiled into a new
DFA $\mathcal{A}_{\Pi'}$ in a background thread within the TEE. To prevent
mid-inference undefined states or execution panics, EHV introduces \textit{Prefix Alignment
Validation}. The background compilation engine evaluates the current active DFA state $q_t$
against the state space of the incoming automaton $\mathcal{A}_{\Pi'}$. The atomic swap of
the active DFA pointer at step $t$ is governed by:

\begin{equation}
\mathcal{A}_{\text{active}} \leftarrow \begin{cases}
\mathcal{A}_{\Pi'} & \text{if } q_t \in \mathcal{A}_{\Pi'} \text{ and } t \in \mathcal{B}, \\
\mathcal{A}_{\text{active}} & \text{otherwise.}
\end{cases}
\end{equation}

where $\mathcal{B}$ denotes the set of transaction boundaries (e.g., between discrete action outputs).
If $q_t \notin \mathcal{A}_{\Pi'}$, the atomic swap is deferred, and a local execution epoch
freeze is maintained until the current transactional boundary completes. This ensures that
(a)~no generation stall occurs mid-sequence, (b)~DFA consistency is maintained within a
single action, and (c)~policy freshness converges to the next action boundary rather than the
next epoch boundary for already-received updates. This state variable has been formally modeled
in the TLA+ specification, verifying safety under arbitrary concurrent interleavings.

\textbf{Grammar Complexity Bound.} The DFA state space grows with policy
complexity. For enterprise deployments with $n$ independent policy constraints,
DFA intersection can produce $\mathcal{O}\left(\prod_i |q_i|\right)$ states in the worst case. The
current architecture assumes domain-specific schemas of \textit{bounded
cardinality}---healthcare clinical action vocabularies typically yield DFAs with
$10^2$--$10^4$ states, well within TEE encrypted memory budgets. For
deployments exceeding this range, hierarchical grammar decomposition (compiling
independent constraint subsets into separate DFAs applied sequentially) is the
recommended scaling strategy. Characterizing the performance cliff under
expanding grammar profiles within TEE encrypted memory constraints is identified
as primary future benchmarking work (Section~\ref{sec:futurework}).

\textbf{Scope Boundary.} GCD enforces \textit{syntactic} compliance: the
generated token sequence conforms to the compiled grammar. Semantic compliance
(whether a syntactically valid action achieves a harmful goal via indirect means)
remains outside the GCD enforcement boundary. A syntactically compliant but
semanticaly harmful action will be executed and signed into the GBOM audit trail
with a clean compliance record; the hardware-rooted enforcement secures the
pipeline mechanics, not the semantic payload. Defense in depth via the ESCALATE
path and external semantic analysis layers is required for high-stakes action
categories. Grammar correctness is a trust dependency: if the policy CFG fails
to capture a relevant constraint, GCD will not enforce it. The grammar
specification process is identified as a primary engineering discipline for EHV
deployments.

Before an agent emits an action $a$, the EHV JIT Compiler evaluates:

\begin{equation}
G(a, C) \in \{\texttt{PERMIT}, \texttt{DENY}, \texttt{ESCALATE}\}
\end{equation}

where $C$ is the current constraint set derived from $S_G$. \texttt{DENY}
routes to a Safe Halt State. \texttt{ESCALATE} triggers a human-in-the-loop
override for non-binary clinical judgment requiring policy interpretation beyond
the compiled grammar.

\begin{figure*}[!t]
\centering
\begin{tikzpicture}[
    node distance=1cm and 1.2cm,
    box/.style={draw, rectangle, align=center, fill=gray!10,
                minimum height=0.8cm, font=\scriptsize},
    tee/.style={draw, rectangle, dashed, inner sep=10pt, fill=blue!5},
    arrow/.style={-{Stealth[scale=1]}, thick}
]
    \node[box, fill=white] (prompt) {User Prompt / Goal};

    \node[box, below=1.5cm of prompt, fill=red!10] (model) {Autoregressive Model\\(Standard GPU)};
    \node[box, right=of model, fill=orange!20] (logits) {Raw Logits};

    \node[box, below=1.5cm of logits, fill=yellow!20] (gcd)
        {GCD Engine\\(DFA State Machine)};
    \node[box, left=of gcd, fill=red!20] (masking)
        {Logits Masking\\($L^{\prime}_t[k]=-\infty$)};

    \node[box, below=of masking, fill=white] (sampler) {Sampler};
    \node[box, right=of sampler, fill=green!20] (token) {Token};
    \node[box, right=of token, fill=blue!20] (gbom)
        {GBOM\\(OSCAL v1.1.2)};

    \node[box, right=3.5cm of prompt, fill=white] (api) {Downstream API};

    \draw[arrow] (prompt) -- (model);
    \draw[arrow] (model) -- (logits);
    \draw[arrow] (logits) -- (gcd)
        node[midway, right, font=\tiny\bfseries\color{red!70!black}] {PCIe DMA};
    \draw[arrow] (gcd) -- (masking)
        node[midway, above, font=\tiny] {Mask Disallowed};
    \draw[arrow] (masking) -- (sampler);
    \draw[arrow] (sampler) -- (token);
    \draw[arrow] (token) -- (gbom)
        node[midway, above, font=\tiny] {Policy-Compliant Token};
    \draw[arrow] (gbom) -- (api)
        node[midway, right, font=\tiny] {Signed / OSCAL-Logged};

    \begin{scope}[on background layer]
        \node[tee,
              fit=(gcd)(masking)(sampler)(token)(gbom),
              label={[font=\bfseries\scriptsize]above:Encrypted CPU Guest Memory (Intel TDX / AMD SEV-SNP)}] (tee_bound) {};
        
        \node[draw, dotted, fit=(gcd)(masking),
              label={[font=\scriptsize]below:JIT PEP (GCD)}] {};

        \node[draw, rectangle, dashed, fill=red!5, inner sep=10pt,
              fit=(model)(logits),
              label={[font=\bfseries\scriptsize]above:Untrusted Accelerator (GPU Memory)}] (gpu_bound) {};
    \end{scope}

\end{tikzpicture}
\caption{System Architecture. The Autoregressive Model runs on an untrusted GPU
accelerator, transferring Raw Logits via PCIe DMA to the JIT PEP executing inside
encrypted CPU guest memory (TEE). The JIT PEP implements Grammar-Constrained Decoding,
masking invalid token scores to $-\infty$ before sampling. Policy-compliant tokens
are then logged to the OSCAL-formatted GBOM.}
\label{fig:architecture}
\end{figure*}
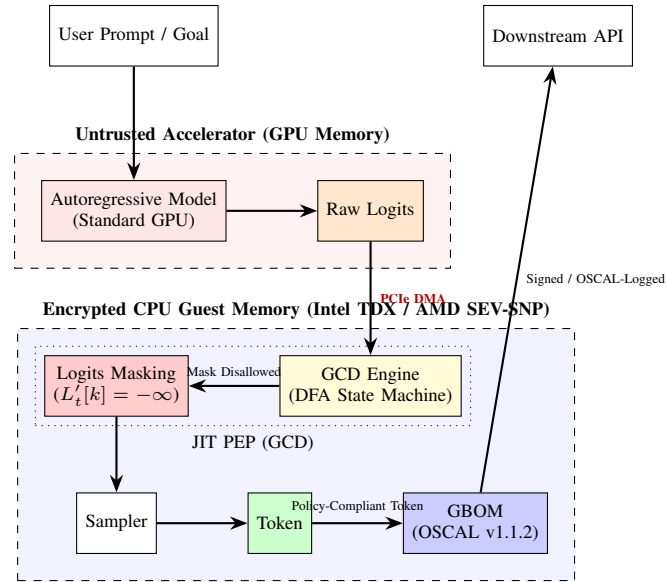

\subsection{Action Schema Extraction Layer (Compatibility Mode)}
\label{subsec:asel}

\textit{(ASEL is retained for backward compatibility with v1.0 deployments
lacking GCD support. In v2, ASEL is not in the primary enforcement critical
path.)}

The PEP can evaluate structured action representations parsed from model outputs
by a pre-PEP Action Schema Extraction Layer (ASEL). ASEL maps unstructured
model output into typed action tuples
$(\texttt{action\_type}, \texttt{parameters}, \texttt{context})$. For example,
the output ``administer 1.5\,mg/m\textsuperscript{2} Vincristine IV'' maps to
$(\texttt{DOSAGE}, \{\texttt{drug}: \texttt{Vincristine}, \texttt{dose}: 1.5,
\texttt{unit}: \texttt{mg/m}^2\}, \texttt{IV})$.

ASEL fidelity is domain-specific and is \textit{not} formally verified within
the current EHV specification. The safety invariant $I_g$ holds conditional on
correct ASEL extraction. Migration to full GCD enforcement (Pillar~3) removes
this unverified dependency.

\subsection{Pillar 4: SPIFFE/SPIRE Workload Identity}
\label{subsec:identity}

In agentic deployments, each agent instance is a Non-Human Identity (NHI) that
must be authenticated before action authorization can be applied. EHV integrates
the SPIFFE (Secure Production Identity Framework For Everyone) standard and
SPIRE runtime~\cite{spiffe} to issue ephemeral, session-bound X.509 SVID
credentials to each agent workload.

The identity model binds:
\begin{itemize}
    \item \textbf{Workload identity}: SPIFFE SVID attesting agent instance
    provenance.
    \item \textbf{Action credential}: RFC~8693 Token
    Exchange~\cite{rfc8693}---scoped, short-lived OAuth~2.0 token authorizing a
    specific action class per epoch.
    \item \textbf{Attestation binding}: The TEE enclave measurement ($\text{MR}$)
    is bound to the action token, ensuring that credentials issued to a valid
    enclave cannot be replayed by a compromised host process.
\end{itemize}

This design supports the goals articulated in the NIST NCCoE concept paper on
AI agent identity and authorization~\cite{nccoe_agent_identity}: separating the
workload identity lifecycle from the policy authorization lifecycle, enabling
fine-grained per-action governance.

\subsection{Separation of Duties and Fault Isolation}
\label{subsec:separation}

EHV separates three functional roles across distinct trust boundaries, aligning
with NIST SP~800-53 AC-5 (Separation of Duties)~\cite{nist_80053}:

\begin{itemize}
    \item \textbf{Policy Administration Point (PAP)}: External to the TEE.
    Human policy authors and automated CI/CD pipelines author, cryptographically
    sign, and publish governance policies. Signing keys are held in external
    Hardware Security Modules (HSMs) or key management services, never inside
    the TEE enclave.
    \item \textbf{Policy Decision/Enforcement Point (PDP/PEP)}: Inside the TEE.
    The JIT PEP verifies policy signatures, merges updates via the CRDT store,
    and enforces the compiled DFA. The PEP cannot author, modify, or override
    policies.
    \item \textbf{Audit Verification Point (AVP)}: External to the TEE. The
    TEE emits cryptographically signed GBOM records
    (Appendix~\ref{app:oscal}). Compliance verification, audit aggregation,
    and regulatory reporting are performed by independent external systems
    consuming these signed records.
\end{itemize}

This separation ensures that a compromised TEE cannot both enforce and audit its
own decisions: a PEP breach results in fail-closed behavior (no actions
permitted), not fraudulent policy injection or audit suppression, because the
PAP signing keys and AVP verification logic reside outside the TEE trust
boundary.

\textbf{Intra-TEE TCB Considerations.} Within the TEE guest memory, the JIT PEP
process consolidates several functions: policy signature verification, CRDT
lattice merging, vector clock state maintenance, CFG-to-DFA compilation, and
logit masking execution. This consolidation expands the intra-TEE Trusted
Computing Base (TCB). If an attacker achieves code execution inside the guest
environment (e.g., via a microarchitectural exploit such as StackWarp), both
policy convergence logic and token filtering are compromised simultaneously.
EHV mitigates this risk through three controls: (1)~the TEE guest image is a
minimal, statically linked binary with no dynamic library loading, reducing the
attack surface; (2)~the PAP signing keys remain external, so a compromised PEP
cannot inject fraudulent policies; and (3)~the AVP independently verifies GBOM
records, so audit integrity is maintained even if enforcement is compromised.
Further TCB decomposition---isolating the CRDT state machine from the DFA
execution engine into separate micro-enclaves---is identified as a hardening
measure for future production deployments.

\section{Formal Verification (TLA+)}
\label{sec:formal}

We specify the EHV system in TLA+ with five state variables:

\begin{itemize}
    \item \texttt{PolicySet}: The current CRDT-merged policy state.
    \item \texttt{AgentAction}: The pending action under evaluation.
    \item \texttt{NetworkState}: $\in \{\texttt{CONNECTED}, \texttt{PARTITIONED}\}$.
    \item \texttt{EnforcementStatus}: $\in \{\texttt{PERMIT}, \texttt{DENY}, \texttt{ESCALATE}\}$.
    \item \texttt{DfaState}: $\in \mathcal{A}_\Pi.\text{States}$ representing the active DFA state $q_t$ tracking token-level prefix alignment.
\end{itemize}

\subsection{Safety Invariant}
\begin{multline}
I_g \triangleq \forall a \in \mathit{AgentAction}:\\
\neg\text{Valid}(a, \mathit{PolicySet}) \implies \mathit{Status} = \texttt{DENY}
\end{multline}

This states that no invalid action can reach a \texttt{PERMIT} state regardless
of the system's execution path.

\subsection{Liveness Property}
\begin{multline}
L \triangleq \forall p \in \mathit{PolicyUpdate}:\\
\Diamond(\mathit{PolicySet}' = \operatorname{merge}(\mathit{PolicySet}, p))
\end{multline}

Every policy update eventually propagates to all nodes (guaranteed by CRDT
convergence).

\subsection{Model Checking Results}
Using the TLC Model Checker (v2026.05.04) with 10 parallel workers, covering all
interleavings of asynchronous policy updates, network partitions, and concurrent
agent actions:

\begin{itemize}
    \item \textbf{States generated}: 1,738
    \item \textbf{Distinct states}: 324
    \item \textbf{State graph depth}: 8
    \item \textbf{Safety violations}: 0
    \item \textbf{Deadlocks}: 0
    \item \textbf{Temporal property violations}: 0 (5 branches checked)
\end{itemize}

The model was configured with $|\texttt{MaxPolicyVersion}| = 5$, $|\texttt{Actions}|
= 3$ (safe\_dosage, unsafe\_dosage, escalate\_case). The invariant $I_g$ held
under all 324 distinct states: non-compliant actions were unreachable in the
verified bounded model within this configuration. The collision probability
across the explored state space is $2.5 \times 10^{-14}$.

The TLA+ specification and TLC output log are provided as supplementary artifacts.%
\footnote{\url{https://github.com/riddhimohansharma/ehv-runtime}}

\textbf{TLA+ Prefix Alignment Verification.} To formally verify that mid-inference
double-buffered DFA updates do not introduce undefined states or execution panics
(Optimization A, Section~\ref{subsec:gcd}), the TLA+ specification was extended
with the \texttt{DfaState} variable and a \texttt{PrefixAligned} invariant. TLC verified
that under all interleavings of background compilation and active token-level swaps,
the active DFA pointer $\mathcal{A}_{\text{active}}$ is only updated when $q_t \in \mathcal{A}_{\Pi'}$.
This guarantees that policy hot-swaps maintain strict syntactic prefix alignment, preventing
state-space out-of-bounds transitions across the entire concurrent lifecycle.

\subsection{Scope and Small-Model Considerations}
\label{subsec:scope}
The model checking results verify the enforcement logic for a bounded
configuration with a single agent variable. Per the small-scope hypothesis~%
\cite{jackson_software}, most design errors in concurrent and distributed systems
manifest in models with small parameter values. We acknowledge the following
scope boundaries:
\begin{itemize}
    \item The specification does not model concurrent multi-agent actions.
    \item Realistic CRDT merge conflict scenarios with vector clocks are not
    modeled.
    \item Unbounded policy version sequences are not covered.
\end{itemize}

Extension to unbounded state spaces via inductive invariants using the TLA+
Proof System (TLAPS) is identified as primary future work. TLAPS would enable
construction of unbounded inductive safety proofs via manual proof obligations;
this requires significant formal methods expertise beyond model checking and is
a multi-month research investment.

\section{Threat Model}
\label{sec:threats}

We enumerate the attack surface specific to governance-compiled agentic systems.

\subsection{Trust Assumptions}
\begin{enumerate}
    \item The TEE hardware root of trust is uncompromised (Intel TDX, AMD
    SEV-SNP at current patch levels).
    \item Policy updates are cryptographically signed by authorized issuers.
    \item The PEP binary within the TEE is measured and attested at epoch
    boundaries.
    \item Network partitions are eventually resolved (partial synchrony model).
    \item Grammar specifications correctly capture the policy intent they
    represent (grammar correctness is a trust dependency, not a formal guarantee).
\end{enumerate}

\subsection{Threat Categories}
\label{subsec:threats}

\begin{table}[htbp]
\caption{EHV Threat Model v2.0}
\centering
\small
\renewcommand{\arraystretch}{1.2}
\begin{tabular}{@{}p{0.6cm}p{2.4cm}p{2.4cm}p{1.0cm}@{}}
\toprule
\textbf{ID} & \textbf{Threat} & \textbf{Mitigation} & \textbf{Residual} \\
\midrule
T1 & Policy poisoning via CRDT injection & Authenticated updates; issuer allowlist; signed DAG mutations; replay protection & Low \\[4pt]
T2 & TEE side-channel (Spectre/Foreshadow/StackWarp) & LFENCE barriers; SMT disabled; small enclave TCB; firmware patch cadence; epoch rotation & Medium \\[4pt]
T3 & Stale epoch exploitation & Configurable $|E_k|$; forced re-attestation on critical updates; bounded epoch TTL & Low \\[4pt]
T4 & Network partition $>$ epoch & CRDT convergence; fail-closed halt state; operator escalation & Low \\[4pt]
T5 & Output obfuscation via adversarial encoding & GCD enforces syntactic grammar constraints; typed action envelopes; note: semantic bypass via syntactically valid payload remains residual & Medium \\[4pt]
T6 & Human override abuse & Signed approval envelopes; RBAC on escalation paths; GBOM logging & Medium \\[4pt]
T7 & Clock skew / NTP poisoning & Vector clock ordering; signed causal DAG; physical clock dependency removed & Low \\
\bottomrule
\end{tabular}
\end{table}

\textbf{T2 Note (StackWarp, 2026):} A newly disclosed architectural
vulnerability class (StackWarp, affecting AMD Zen~1 through Zen~5 processors)
can manipulate the stack pointer inside an SEV-SNP guest and break guest
integrity~\cite{stackwarp}. Mitigation requires firmware-level patches in
addition to the LFENCE and SMT-disable mitigations. EHV deployments on AMD
SEV-SNP must maintain a current firmware patch schedule as a non-optional
security control.

\textbf{T2 Note (Performance Impact):} The performance impact of host-side TEE
mitigations (LFENCE barriers, SMT disable) applies to CPU-bound operations.
The GCD logit masking step---the latency-critical enforcement operation---executes
as a vectorized array operation on the GPU via the LogitsProcessor interface,
outside the CPU mitigation boundary. LLM inference latency is dominated by the
GPU forward pass (10--30ms per token); the GCD mask application adds
$\sim$0.033ms per step. However, logit buffers must transit between host CPU
memory and GPU memory via PCIe DMA transfers; if host-side LFENCE serialization
stalls amplify the synchronization latency between token generation steps, the
effective per-step overhead may increase. EHV mitigates this PCIe DMA bottleneck via
\textit{Asynchronous Mask Pre-fetching}. Rather than waiting for the raw logits to
be generated on the GPU, transferred to the CPU TEE, masked, and returned, the CPU TEE
pre-calculates the valid token transition mask $\Sigma_{t+1}$ based on predicted DFA
state transitions and pushes it to a secure GPU memory buffer \textit{in parallel} with
the GPU's current token forward pass. This hides the PCIe transfer and LFENCE serialization
penalties entirely within the GPU compute latency envelope, maintaining sub-millisecond
enforcement without pipeline stalls. In practice, PCIe DMA transfers are
bus-bandwidth-bound rather than CPU-pipeline-bound, limiting this amplification
effect. Empirical measurement of the combined TEE mitigation + DMA
synchronization overhead on production hardware is required
(Section~\ref{sec:futurework}).

\textbf{T5 Note (GCD Scope):} GCD enforces \textit{syntactic} compliance.
An adversary who constructs a syntactically valid but semantically harmful
action (e.g., a valid dosage token within permitted range that achieves harm
through drug interaction context) is not blocked by GCD alone. Defense in depth
via human clinical oversight (ESCALATE path) and semantic intent analysis is
required for high-stakes action categories.

\begin{figure}[!t]
\centering
\resizebox{\columnwidth}{!}{%
\begin{tikzpicture}[
    node distance=1.35cm and 1.35cm,
    nodebox/.style={draw, rectangle, fill=gray!10, minimum height=0.6cm,
                    text width=2.15cm, align=center, font=\scriptsize},
    arrow/.style={-{Stealth[scale=1]}, thick}
]
    \node[nodebox] (pap) {Policy Admin Point\\(PAP)\\$V_P = [1,0,0]$};
    \node[nodebox, below left=of pap] (pep1) {JIT PEP 1\\(TEE)\\$V_1 = [1,1,0]$};
    \node[nodebox, below right=of pap] (pep2) {JIT PEP 2\\(TEE)\\$V_2 = [1,0,1]$};

    \draw[arrow] (pap) -- (pep1)
        node[midway, above, sloped, font=\tiny] {Publish $p_1$};
    \draw[arrow] (pap) -- (pep2)
        node[midway, above, sloped, font=\tiny] {Publish $p_1$};
    \draw[arrow, <->] (pep1) to[bend left=12]
        node[midway, above, font=\tiny] {CRDT Merge $\sqcup$} (pep2);
    \draw[arrow, dashed] (pep1) to[bend right=12]
        node[midway, below, font=\tiny] {Merge: $\max(V_1,V_2)$} (pep2);

\end{tikzpicture}
}
\caption{Causal Graph CRDT Policy Synchronization. Policy updates are propagated
across the distributed network using Vector Clocks (Lamport causal ordering) and
merged via a Join-Semilattice $\sqcup$, removing dependencies on vulnerable physical
clocks (NTP).}
\label{fig:causal}
\end{figure}
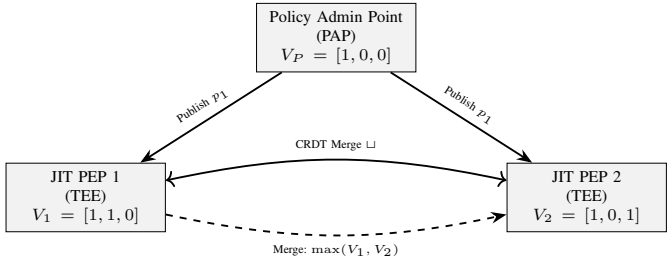

\subsection{Failure Mode: Non-TEE Environments}
In environments lacking Confidential Computing support, the Sub-millisecond
enforcement target degrades. The system falls back to out-of-band audit with
$\mathit{GL} > 0$, governed by NIST SP~800-53 SI-17 fail-safe provisions~\cite{nist_80053}.
This is the primary architectural friction point for legacy infrastructure
deployments.

\subsection{Epoch Staleness Window (ESW) Analysis}
Within a policy epoch $E_k$, a critical policy update arriving at time $t_u$
is not enforced until the next epoch boundary. The maximum staleness window is:

\begin{equation}
\mathit{ESW}_{\max} = |E_k| - 1
\end{equation}

For healthcare parameters ($\lambda = 500{,}000$ actions/hour aggregate,
$|E_k| = 60\text{s}$):

\begin{multline}
N_{\text{stale}} \leq \lambda \cdot \mathit{ESW}_{\max}\\
= \frac{500{,}000}{3{,}600} \times 59 \approx 8{,}194 \text{ actions}
\end{multline}

Compared to the legacy $N_{\text{unsafe}} = 168{,}000{,}000$ under 14-day $\mathit{GL}$,
this represents a $\mathbf{5}$ \textbf{order-of-magnitude reduction}. For
ultra-critical updates (e.g., drug withdrawal signals), EHV supports forced
mid-epoch re-attestation via an \texttt{EMERGENCY\_EPOCH\_RESET} signal, reducing
$\mathit{ESW}$ to network propagation latency ($<1$s). The emergency reset channel itself
introduces an attack surface: an adversary who can trigger spurious resets
creates a Denial of Service (DoS) via continuous re-attestation overhead, while
an adversary who can block the signal preserves the staleness window. EHV
mitigates reset-channel abuse through cryptographic authentication of reset
signals (requiring PAP-level signing authority) and rate-limiting reset
acceptance to at most one forced reset per $|E_k|/2$ interval.

\subsection{Fail-Closed Partition Semantics}
When a network partition persists beyond the epoch boundary $|E_k|$ and the
local node cannot reach the attestation authority, EHV enforces strict
fail-closed semantics:

\begin{equation}
t - t_{\text{last\_attest}} > |E_k| \implies G(a, C) = \texttt{DENY} \quad \forall a
\end{equation}

The JIT PEP transitions to a \texttt{Safe Halt State}, blocking all outgoing
tool and action executions until attestation is restored. This guarantees that
no agent executes under an indefinitely stale policy state during extended
network outages.

\textbf{Availability Attack Surface.} The fail-closed design creates an
adversary-exploitable availability vulnerability: an external attacker who can
sustain a network partition (e.g., via targeted network jamming or DNS
poisoning against the attestation authority) can force all agents in the
affected zone into continuous halt states, crippling operational availability.
For safety-critical deployments (e.g., medical monitoring, energy grid control),
this trade-off is explicit: EHV prioritizes safety (no action under stale
policy) over availability (continuous operation under uncertainty). Deployments
requiring high availability under adversarial network conditions should implement
redundant attestation authority paths and geographic failover for the
attestation endpoint, reducing the single-path partition vulnerability.

\section{Case Study: Pediatric Oncology Dosage}
\label{sec:casestudy}

Consider an FDA-mandated reduction in Vincristine dosage from
1.5\,mg/m\textsuperscript{2} to 0.75\,mg/m\textsuperscript{2} due to a new
neurotoxicity signal.

\subsection{Legacy System (GL = 14 days)}
Manual protocol review, committee approval, EHR update, and staff retraining
yield $\mathit{GL} = 14$ days. During this interval, across 5,000 Physician Twin
instances processing 100 recommendations/hour:

\begin{multline}
N_{\text{toxic}} = 5{,}000 \times 100 \times 24 \times 14\\
= 168{,}000{,}000 \text{ actions under stale policy}
\end{multline}

Even at a conservative 0.03\% violation rate, this produces 50,400 potentially
toxic dosage recommendations before the policy is enforced.

\subsection{EHV System (GL < 1ms, target)}
The dosage update propagates via Causal CRDT in $<$1 second. The compiled GCD
grammar for the Vincristine dosage policy sets the maximum permitted token to
\texttt{0.75}; within the scope of the compiled grammar, any model output
exceeding this value is excluded from the token probability mass before sampling
(the logit is set to $-\infty$, yielding zero probability after softmax). The
action cannot be emitted regardless of the underlying LLM's training
distribution. This guarantee is syntactic: it holds for constraints expressible
in the policy grammar but does not extend to semantic bypass vectors outside
the grammar's scope (Section~\ref{subsec:gcd}).

The GBOM (Section~\ref{subsec:gbom}) issues a cryptographic receipt binding each
clinical recommendation to the specific policy version (Merkle root $H_{\mathrm{p}}$), TEE
attestation epoch, and enforcement outcome, enabling post-hoc verification of
any recommendation's compliance provenance.

\section{Discussion}
\label{sec:discussion}

\subsection{The Velocity-Ethics Co-Production Principle}
In EHV-compliant systems, deployment velocity $V$ and governance integrity $I$
are positively correlated:

\begin{equation}
\frac{\partial V}{\partial I} \geq 0 \quad \text{(EHV)} \quad \text{vs.} \quad
\frac{\partial V}{\partial I} < 0 \quad \text{(traditional)}
\end{equation}

The causal mechanism underlying this sign reversal is \textit{pre-clearance
elimination of post-hoc audit}. In traditional architectures, governance imposes
friction as a linear cost: $V = V_0 - k \cdot I$, where $k$ represents audit
overhead per unit of integrity. In EHV, because compliance is verified at
inference time, no post-deployment audit backlog accumulates. Formally:
$V_{\text{EHV}} = V_0 \cdot f(I)$ where $f(I) \geq 1$ when pre-execution
verification reduces the need for retrospective compliance gates. Within the EHV
enforcement model, governance becomes the mechanism of acceleration rather than
friction. This co-production property is conditional on the assumptions in
Section~\ref{sec:threats}: TEE integrity, grammar correctness, and eventual CRDT
convergence. If any assumption is violated, the system fails closed rather than
producing a false co-production signal.

\subsection{Governance Bill of Materials (GBOM) for M\&A Due Diligence}
\label{subsec:gbom}
EHV introduces the \textbf{Governance Bill of Materials (GBOM)}: a cryptographic
audit trail that binds each autonomous decision to the specific policy version
(Merkle root $H_{\mathrm{p}}$), TEE attestation epoch, GCD DFA state, and enforcement
outcome that governed it. The GBOM is formatted as OSCAL Assessment Results
(v1.1.2)~\cite{oscal_ref,oscal_ai_2026}, the NIST-developed schema used in
FedRAMP compliance automation. This enables acquirers in M\&A due diligence to
verify the governance posture of an AI stack with the same rigor applied to
financial audits.
See Appendix~\ref{app:oscal} for a representative OSCAL GBOM sample.

\subsection{Regulatory Alignment}
\label{subsec:regulatory}

Table~\ref{tab:controls} maps EHV controls to applicable regulatory and
standards requirements. EHV \textit{supports compliance evidence generation}
for these frameworks; it does not replace the organizational controls, legal
review, or conformity assessment required for legal compliance.

\begin{table}[htbp]
\caption{Controls-to-Requirements Traceability Matrix}
\label{tab:controls}
\centering
\small
\renewcommand{\arraystretch}{1.2}
\begin{tabular}{@{}p{2.1cm}p{2.0cm}p{2.2cm}@{}}
\toprule
\textbf{Requirement} & \textbf{EHV Control} & \textbf{Evidence Artifact} \\
\midrule
NIST AI RMF Govern~1.2 (Accountability) & Signed policy lifecycle; audit trail; incident hooks & GBOM; policy mutation log \\[4pt]
NIST AI RMF Manage~2.4 (Monitoring) & GCD enforcement + epoch-scoped continuous attestation & Per-action OSCAL observation records \\[4pt]
NIST SP~800-207 \S3 (PEP/PE/PA) & Identity + per-action verification; SPIFFE SVID credentials; JIT PEP maps to ZTA PEP & Attestation log; token exchange records \\[4pt]
NIST SP~800-53 AC-5 (Separation of Duties) & PAP/PDP/PEP separation across trust boundaries (Section~\ref{subsec:separation}) & Signing key isolation; GBOM independence \\[4pt]
EU AI Act Article~12 (Logging) & Per-action signed records; policy version binding & OSCAL assessment-results \\[4pt]
FDA PCCP~\cite{fda_pccp} & Constraint-set updateable without redeployment via epoch mechanism & Policy epoch transition log \\[4pt]
Human oversight & ESCALATE state with signed approval envelope & Signed override record; GBOM entry \\[4pt]
Change control & Epoch update discipline; Merkle-rooted policy commits & Policy mutation log \\
\bottomrule
\end{tabular}
\end{table}

\subsection{Deployment Topology and Operational Considerations}
\label{subsec:deployment}
EHV is designed for \textit{high-stakes regulated verticals}---healthcare
clinical agents, financial compliance automation, critical infrastructure
control---where the cost of a single unsafe autonomous action exceeds the
infrastructure premium of Confidential Computing. It is not positioned as a
general-purpose middleware for consumer-facing chatbots or advisory systems.

Confidential Computing infrastructure is available today on all three major
clouds: AWS (\texttt{c6a.metal} with SEV-SNP), GCP (\texttt{C3D} with SEV-SNP),
and Azure (\texttt{DCasv5} with SEV-SNP/TDX). The cost premium is 10--20\%
over standard instances for equivalent compute. SPIRE federation enables
regional TEE clusters with automatic credential rotation, reducing global
synchronization complexity to a SPIRE server mesh operating alongside the
CRDT policy store.

\subsection{TEE Portability}
\label{subsec:tee_portability}
The EHV software architecture---GCD engine, CRDT policy store, SPIFFE identity
layer, GBOM logging---is specified independently of any TEE vendor. The TEE
provides two abstract interfaces: (1)~encrypted memory isolation and
(2)~remote attestation. These interfaces are implemented by AMD SEV-SNP,
Intel TDX, ARM CCA, and NVIDIA H100 Confidential Computing. EHV deployments
can migrate between TEE backends without modifying the policy compilation,
enforcement logic, or audit format.

EHV does require \textit{some} Confidential Computing substrate---it is not
hardware-agnostic in the sense of running on commodity hosts. The architecture
is TEE-vendor-portable, not TEE-optional. Non-TEE deployments degrade to
out-of-band auditing (Limitation~1 below).

\subsection{Comparison: Prompt-Layer Policy-as-Code}
\label{subsec:pac_comparison}
An alternative to EHV's approach is Declarative Policy-as-Code (PaC) enforced
via system-prompt engineering, with compliance verified through transparency
ledgers. Table~\ref{tab:pac_compare} compares these architectures.

\begin{table}[htbp]
\caption{Architectural Comparison: PaC vs.\ EHV}
\label{tab:pac_compare}
\centering
\small
\renewcommand{\arraystretch}{1.2}
\begin{tabular}{@{}p{1.8cm}p{2.5cm}p{2.5cm}@{}}
\toprule
\textbf{Property} & \textbf{System-Prompt PaC + Ledger} & \textbf{EHV (GCD + TEE)} \\
\midrule
Jailbreak resistance & Low---system prompts are bypassable via adversarial inputs & High---DFA rejects invalid tokens before sampling \\[4pt]
Enforcement timing & Post-hoc (logged after execution) & Pre-execution (blocked at token level) \\[4pt]
Security guarantee & Probabilistic (LLM may ignore prompt) & Deterministic syntactic constraint (within grammar scope) \\[4pt]
Semantic bypass & Vulnerable & Equally vulnerable \\[4pt]
Hardware cost & None & 10--20\% CVM premium \\[4pt]
Deployment complexity & Low & High \\
\bottomrule
\end{tabular}
\end{table}

PaC is appropriate for low-stakes advisory systems where post-hoc accountability
suffices. For regulated domains where a single unconstrained action carries
regulatory liability (e.g., clinical dosage, financial trade execution), the
pre-execution guarantee of GCD+TEE justifies the infrastructure premium.

\subsection{Limitations and Future Empirical Work}
\begin{enumerate}
    \item \textbf{TEE dependency}: Confidential Computing hardware is required
    for the full enforcement model. Non-TEE deployments degrade to out-of-band
    auditing.
    \item \textbf{Epoch granularity}: The trade-off between freshness and
    attestation cost requires domain-specific tuning.
    \item \textbf{Grammar correctness}: The GCD trust boundary shifts from ASEL
    parsing fidelity to grammar specification correctness. Formal verification
    of grammar completeness is future work.
    \item \textbf{Bounded model checking}: The TLA+ verification covers a bounded
    configuration (depth~8, 324 distinct states). Extension to unbounded state
    spaces via TLAPS inductive invariants is primary future work.
    \item \textbf{No empirical latency data}: The $<1$ms enforcement latency
    target is architectural. Empirical validation on production SEV-SNP hardware
    is required before this claim can be substantiated in a peer-reviewed venue.
    \item \textbf{Intra-TEE TCB breadth}: The JIT PEP consolidates CRDT
    merging, DFA compilation, and logit masking within a single TEE process
    boundary. Decomposition into isolated micro-enclaves is future hardening
    work.
    \item \textbf{Grammar complexity scaling}: DFA state explosion under
    large-cardinality policy sets within TEE encrypted memory (where paging
    incurs cryptographic overhead) may introduce performance cliffs.
    Hierarchical grammar decomposition and empirical memory profiling are
    required.
    \item \textbf{Fail-closed availability}: The strict CP design creates an
    adversary-exploitable availability attack surface via sustained network
    partitioning. Redundant attestation authority paths mitigate but do not
    eliminate this risk.
\end{enumerate}

\section{Empirical Validation Plan}
\label{sec:futurework}

The following benchmark plan defines the measurement protocol for validating
EHV's latency claims. These benchmarks constitute primary future work.

\textbf{Target Platform}: AMD SEV-SNP (AWS, GCP, or Azure confidential VMs).
AMD SEV-SNP is selected as the primary MVP target for lower cloud setup friction
and broader instance availability relative to Intel TDX~\cite{amd_sev}.

\textbf{Workload Matrix}:
\begin{itemize}
    \item W1: Safe action permitted (PERMIT baseline).
    \item W2: Unsafe dosage DENY (GCD masking active).
    \item W3: Ambiguous clinical case ESCALATE.
    \item W4: Mid-epoch policy update (boundary behavior).
    \item W5: Network partition past epoch (fail-closed transition).
\end{itemize}

\textbf{Metrics}: Per-token latency (baseline vs.\ GCD vs.\ GCD+attestation);
enclave enter/exit overhead; policy propagation latency; fail-closed transition
time.

\textbf{Reproducibility}: Fixed random seed; pinned model version (1B--7B
parameter range); frozen policy corpus; recorded instance type, kernel version,
SEV-SNP firmware revision, enclave measurement, and policy Merkle root in every
row. Hyperthreading disabled to reduce side-channel noise.

Expected overhead based on published SEV-SNP literature: 2--10\% for general
workloads; $<$20\% for LLM inference; remote attestation cost amortized to
$<1$\,ms per inference via epoch caching~\cite{amd_sev}.

\section{Conclusion}
\label{sec:conclusion}

EHV transforms governance from a manual gate into a hardware-rooted system
invariant. By elevating Grammar-Constrained Decoding to the primary Policy
Enforcement Point within the token-generation pipeline, backed by Causal
CRDT-synchronized policy state and TEE-anchored attestation caching, EHV reduces
Governance Latency by orders of magnitude for agentic systems in regulated
domains.

The bounded formal verification demonstrates that non-compliant actions were
unreachable in the verified bounded state space under all explored interleavings.
The Velocity-Ethics Co-Production Principle establishes that governance integrity
and deployment speed are not adversarial but co-productive when enforcement is
architectural rather than procedural.

EHV's differentiating contribution is the integration of four independently
validated mechanisms---GCD, Causal CRDT, TEE attestation caching, and bounded
formal verification---into a single, hardware-rooted enforcement architecture
for regulated agentic systems. No contemporaneous system achieves this four-way
integration.

Future work includes:
\begin{enumerate}
    \item Extending the TLA+ specification to unbounded state spaces via TLAPS
    inductive invariants.
    \item Benchmarking enforcement latency on production TEE hardware (AMD
    SEV-SNP, Intel TDX) per the plan in Section~\ref{sec:futurework}.
    \item Formal grammar completeness verification for healthcare-specific
    clinical action schemas.
    \item CRDT propagation latency characterization under realistic WAN
    conditions.
\end{enumerate}

\section*{AI Tool Disclosure}
AI tools (Claude, Gemini) were used for prose refinement, LaTeX formatting,
and literature survey assistance. All technical claims, formal specifications
(TLA+), architectural design decisions, mathematical derivations, and the system
architecture are the original intellectual work of the author. The TLA+
specification was authored and independently verified by the researcher using
the TLC model checker. The proof-of-concept implementation was developed by
the author.

\section*{Acknowledgments}
The author acknowledges the NIST CAISI working group whose publications
informed the identity model in Section~\ref{subsec:identity}.


\appendix

\section{Grammar-Constrained Decoding: Pseudocode and Integration}
\label{app:gcd}

\subsection{DFA Compilation}
Given a policy $\Pi$ expressed as a Context-Free Grammar (CFG) over the action
vocabulary, we compile $\Pi$ to a DFA $\mathcal{A}_\Pi$ using standard
subset-construction. For efficiency, we represent $\mathcal{A}_\Pi$ as a
Compressed Sparse Row (CSR) matrix over the token vocabulary, enabling
$\mathcal{O}(1)$ transition lookup per token.

\subsection{Logits Masking Pseudocode}
\begin{algorithmic}
\REQUIRE prefix tokens $\tau$, raw logits $L_t$, DFA $\mathcal{A}_\Pi$
\STATE $q \leftarrow \mathcal{A}_\Pi.\text{start}$
\FOR{$t \in \tau$}
    \STATE $q \leftarrow \mathcal{A}_\Pi.\text{transition}(q, t)$
    \IF{$q = \emptyset$}
        \RETURN \texttt{DENY}
    \ENDIF
\ENDFOR
\STATE $\Sigma_t \leftarrow \mathcal{A}_\Pi.\text{allowed}(q)$
\FOR{$k \in \text{Vocabulary}$}
    \IF{$k \notin \Sigma_t$}
        \STATE $L_t[k] \leftarrow -\infty$
    \ENDIF
\ENDFOR
\RETURN $L_t$
\end{algorithmic}

\subsection{Integration Rules}
\begin{itemize}
    \item Masking MUST occur before softmax at every generation step.
    \item The DFA state $q_t$ and policy Merkle root $H_{\mathrm{p}}$ are logged to the
    GBOM for each token.
    \item Action outputs are wrapped in typed envelopes before downstream tool
    execution.
    \item The GCD engine runs within encrypted guest memory; the DFA state is
    sealed by the TEE.
\end{itemize}


\section{OSCAL GBOM Sample (v1.1.2)}
\label{app:oscal}

The following is a representative OSCAL~1.1.2 Assessment Results record
binding a single clinical action to its governance provenance:

\begin{lstlisting}[basicstyle=\tiny\ttfamily,
  breaklines=true, frame=single]
{
  "$schema":
    "https://pages.nist.gov/OSCAL/schemas/
     json/1.1.2/oscal_assessment-results_schema.json",
  "assessment-results": {
    "uuid": "f81d4fae-7dec-11d0-a765-00a0c91e6bf6",
    "metadata": {
      "title": "EHV Runtime Governance Bill of Materials",
      "last-modified": "2026-05-27T00:00:00Z",
      "version": "2.0.0",
      "oscal-version": "1.1.2"
    },
    "results": [
      {
        "uuid": "a1b2c3d4-e5f6-7890-abcd-ef1234567890",
        "title": "Dosage Recommendation Action - Epoch E-042",
        "description": "Vincristine dosage recommendation",
        "start": "2026-05-27T00:00:00Z",
        "end":   "2026-05-27T00:00:00.001Z",
        "reviewed-controls": {
          "control-selections": [
            {"include-controls": [{"control-id": "si-17"}]}
          ]
        },
        "observations": [
          {
            "uuid": "obs-001",
            "description": "GCD enforcement outcome",
            "props": [
              {"name": "policy_merkle_root",
               "value": "sha256:abc123..."},
              {"name": "tee_measurement",
               "value": "sevsnp:mrenclave:def456..."},
              {"name": "epoch_id",      "value": "E-042"},
              {"name": "dfa_state",     "value": "q17"},
              {"name": "enforcement",   "value": "PERMIT"},
              {"name": "spiffe_svid",
               "value": "spiffe://ehv.example/agent/twin-001"}
            ]
          }
        ],
        "findings": [
          {
            "uuid": "find-001",
            "title": "Action compliant with epoch policy",
            "description":
              "Dosage within GCD-enforced range 0.0-0.75 mg/m2",
            "target": {
              "type": "objective-id",
              "target-id": "ehv-si-17",
              "status": {"state": "satisfied"}
            }
          }
        ]
      }
    ]
  }
}
\end{lstlisting}


\end{document}